\documentclass[letterpaper]{article} 
\usepackage{aaai2027} 
\usepackage[hyphens]{url}  
\usepackage{graphicx} 
\usepackage{natbib}  
\usepackage{caption} 
\usepackage{algorithm}
\usepackage{algorithmic}
\usepackage{amsmath}
\usepackage{amssymb}
\usepackage{multirow}
\usepackage{newfloat}
\usepackage{listings}
\DeclareCaptionStyle{ruled}{labelfont=normalfont,labelsep=colon,strut=off} 
\floatstyle{ruled}
\newfloat{listing}{tb}{lst}{}
\floatname{listing}{Listing}

\usepackage{booktabs}

\title{TexTailor: Texture-Preserving Video Virtual Try-On \\
via Adaptive Garment Conditioning}

\author{
Zijing Qin\textsuperscript{\rm 1}\thanks{Equal contribution.},
Jun Zhou\textsuperscript{\rm 1}\footnotemark[1],
Ruicheng Zhang\textsuperscript{\rm 1},
Jiaqi Hou\textsuperscript{\rm 1},
Zunnan Xu\textsuperscript{\rm 1},
\\
Ronghui Li\textsuperscript{\rm 1},
Zhenyu Xie\textsuperscript{\rm 2},
Xiu Li\textsuperscript{\rm 1}\thanks{Corresponding author.}
}
\affiliations{
\textsuperscript{\rm 1}Tsinghua University, China\\
\textsuperscript{\rm 2}Mohamed bin Zayed University of Artificial Intelligence, UAE\\
}

\begin{document}
\maketitle

\begin{abstract}

Video virtual try-on has attracted increasing attention due to its broad potential in digital fashion and intelligent e-commerce. 
However, existing methods primarily focus on low-resolution settings and still face substantial challenges when extended to high-resolution scenarios. 
These limitations can be attributed to two main factors: (1) the insufficient utilization of rich garment reference information, and (2) the lack of explicit positional modeling between garment and video representations during cross-modal interaction, which weakens fine-grained local correspondence.
To address these issues, we propose TexTailor, a high-fidelity video virtual try-on framework built upon a pretrained video Diffusion Transformer. 
Specifically, we introduce a timestep-adaptive modulation mechanism to dynamically adjust garment visual representations throughout denoising. 
We further develop a frame-aligned positional encoding strategy to strengthen garment-to-video correspondence, together with a multi-source injection design that reduces interference among heterogeneous conditions.
Extensive experiments on multiple video virtual try-on benchmarks, including the high-resolution Eevee dataset, demonstrate that TexTailor achieves competitive performance in garment detail preservation, temporal consistency, and overall video quality.

\end{abstract}

\begin{figure*}[t]
    \centering
    \includegraphics[width=\textwidth]{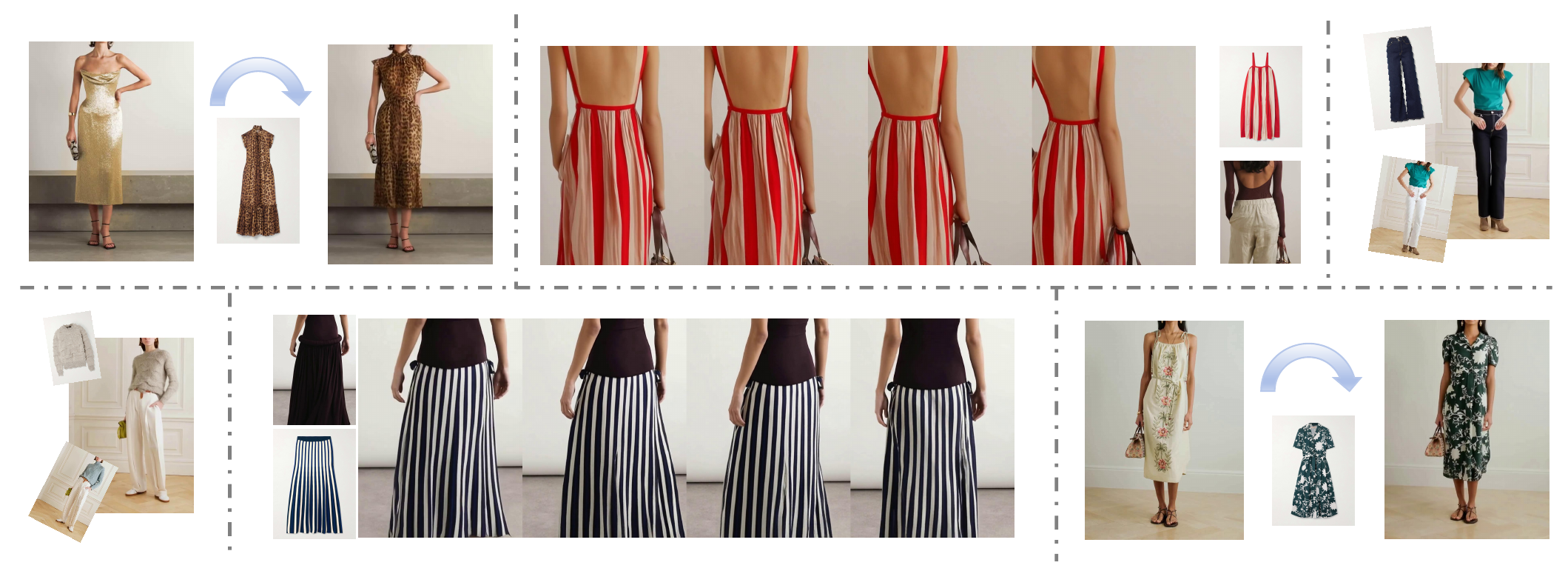}
    \caption{
    TexTailor generates high-fidelity video virtual try-on results
    with temporally consistent motion while preserving garment
    textures, patterns, and local structures across diverse poses
    and viewpoints.
    }
    \label{fig:teaser}
\end{figure*}

\begin{figure*}[t]
    \centering
    \includegraphics[width=\textwidth]{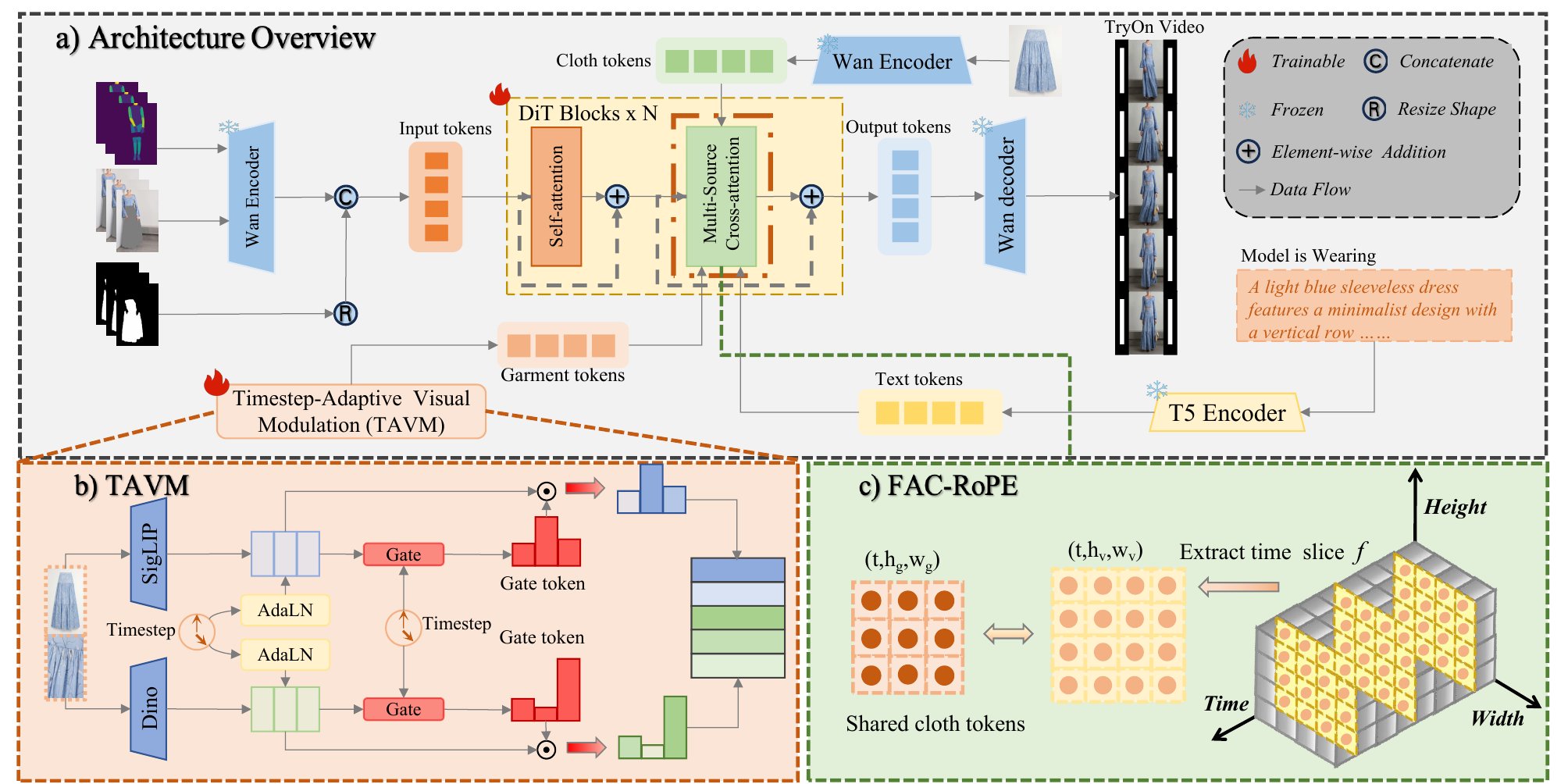}
    \caption{Overview of TexTailor. (a) A DiT backbone integrates structural input, text token, visual token, and garment latent token through MCAI for video try-on generation. (b) TAVM modulates visual features with timestep-adaptive token-wise gating. (c) FAC-RoPE assigns frame-aligned temporal coordinates to static garment latent tokens, preserving native 3D RoPE while establishing stable garment-to-video spatial correspondence.
}
    \label{fig:overview}
\end{figure*}

\section{Introduction}

Recent advances in image generation have enabled high-fidelity and controllable visual synthesis~\cite{zhou2025fireedit,huang2026consistentid,hu2026embedding,liu2025controllable}, while video generation further extends these capabilities to dynamic content with temporally coherent motion~\cite{kong2024hunyuanvideo,xu2025hunyuanportrait,zhang2026zo3t,zhang2026kvpo,hu2026physeditworld}. Building upon these developments, virtual try-on has progressed from static image synthesis~\cite{wang2024stablegarment,choi2024improving,chong2024catvton,xu2025ootdiffusion,zhang2024mmtryon,yang2025omnivton,guo2025any2anytryon} toward video-based scenarios~\cite{chong2025catv2ton,fang2024vivid,karras2024fashion,li2025magictryon,zheng2024viton,zuo2025dreamvvt}, where garments must remain faithful under continuous human motion and viewpoint changes. Unlike image-based try-on, video virtual try-on requires not only preserving fine-grained garment appearance, including shapes, textures, and patterns, but also maintaining temporal consistency of human identity and motion. Balancing garment fidelity with coherent motion therefore remains a fundamental challenge.

Despite promising progress on benchmarks such as ViViD~\cite{fang2024vivid} and VVT~\cite{dong2019fw}, existing video virtual try-on methods are mostly developed on relatively low-resolution data, where fine-grained garment fidelity is less emphasized. At higher resolutions, garment details such as textures, stitching, patterns, and subtle material structures become more demanding to preserve and remain susceptible to blurring, loss, and temporal drift, as further revealed by recent high-resolution benchmarks such as Eevee ~\cite{zeng2025eeveecloseuphighresolutionvideobased}. This calls for more effective exploitation of fine-grained garment conditions.
Beyond limited training resolution, existing methods often make insufficient use of garment conditions. Garment visual features are commonly injected in a fixed manner throughout denoising, rather than being dynamically utilized at different denoising stages. In addition, existing methods often perform garment-latent cross-attention without explicitly modeling positional correspondence between the garment and video latent spaces, weakening the consistency of local garment-to-video matching. Moreover, heterogeneous conditions, including text, visual feature, and garment latent, are often fused within a shared attention space, potentially causing interference and limiting the effectiveness of garment-specific guidance during high-resolution video generation.

To address these limitations, we propose TexTailor, a high-fidelity video virtual try-on framework built upon a pretrained video Diffusion Transformer. Our framework improves garment conditioning through stage-adaptive visual modulation, explicit frame-wise garment-to-video alignment, and decoupled multi-source injection.
First, Timestep-Adaptive Visual Modulation (TAVM) combines timestep-conditioned AdaLN with timestep-aware token-wise gating to emphasize visual representation relevant to each denoising stage. This enables stage-adaptive garment guidance from global structural cues at early stage to fine-grained local details at later stage.
Second, Frame-Aligned 3D Garment Cross-RoPE (FAC-RoPE) assigns garment key the temporal coordinate of the corresponding video frame while retaining the native 3D RoPE of the video query. This avoids artificial temporal offsets and establishes stable patch-level garment-to-video correspondence.
Finally, the Multi-Source Cross-Attention Injection (MCAI) module decouples the conditioning process by processing textual cues, timestep-modulated visual features, and frame-aligned garment latents through parallel cross-attention pathways. By isolating interactions among these heterogeneous conditioning sources, MCAI effectively mitigates cross-modal interference while harmonizing their complementary information, ultimately enabling more faithful garment appearance retention and higher-fidelity video virtual try-on.

We conduct extensive experiments on multiple video virtual try-on benchmarks, including the high-resolution Eevee benchmark. The results demonstrate that TexTailor achieves competitive performance in garment detail preservation, temporal consistency, and overall video quality.

Our contributions can be summarized as follows:
\begin{itemize}
    \item We propose TexTailor, a high-fidelity video virtual try-on framework that effectively preserves both global garment appearance and fine-grained local details while maintaining temporal consistency across generated videos.

\item We propose a dynamic garment conditioning mechanism that calibrates garment feature injection strength across denoising timesteps while integrating customized 3D positional encodings to guarantee rigorous garment-to-video alignment, thereby ensuring precise and continuous preservation of complex garment textures.

    \item Extensive experiments on multiple video virtual try-on benchmarks demonstrate that TexTailor consistently improves garment fidelity, temporal consistency, and high-resolution generation quality compared with existing approaches.
\end{itemize}

\section{Related Work}
\subsection{Video Generation}

Recent advances in diffusion models and Transformers have led to significant progress in video generation. 
Representative approaches such as AnimateDiff~\cite{guo2023animatediff}, Tune-A-Video \cite{wu2023tune}, Dreamix~\cite{molad2023dreamix}, Make-A-Video \cite{singer2022make}, and KVPO~\cite{zhang2026kvpo} have demonstrated strong capabilities in generating temporally coherent and visually realistic video sequences by modeling both spatial appearance and motion dynamics.
More recent methods further improve video realism through stronger temporal attention, reference-guided generation, or motion-specific modules, enabling more stable synthesis across frames~\cite{lai2025tracktention,deng2025magref,hong2025audio,zhang2025mind,hu2026identity}.

Despite these achievements, video virtual try-on remains substantially more challenging than general-purpose video generation~\cite{guo2023animatediff}. In addition to maintaining temporal consistency, a try-on model must faithfully preserve garment-specific attributes, such as texture, patterns, material properties, and fine structural details, while also ensuring plausible interaction between clothing and human motion. Therefore, directly applying generic video generation models to virtual try-on is often insufficient, and task-specific garment modeling and conditioning mechanisms are required.

\subsection{Video Virtual Try-On}

Compared with image-based virtual try-on \cite{zhu2023tryondiffusion,gou2023taming,cui2025street,kim2024stableviton,shim2024towards,zhou2025learning,sai2026modaflow}, video virtual try-on provides a more realistic and practical user experience by presenting garments under dynamic motion and changing viewpoints \cite{chen2025dress,wei20253dv,kang2024mirror,pan2025once,zuo2025dreamvvt,li2025pursuing,zheng2024dynamic,chang2025pemf}. 
Early video virtual try-on methods mainly focused on improving temporal coherence through warping, blending, and optical-flow-based smoothing \cite{dong2019towards}. 
With the rapid development of diffusion-based generative models, recent approaches have increasingly adopted DiT-based architectures for this task \cite{chong2025catv2ton,chen2025dress,li2025magictryon}. 
ViViD \cite{fang2024vivid} extends image diffusion models to video try-on by introducing temporal modeling modules and garment encoding. 
WildVidFit \cite{he2024wildvidfit} employs controllable diffusion to improve video try-on generation, while RealVVT \cite{li2025realvvt} emphasizes photorealistic and temporally stable synthesis in dynamic scenes. 
More recently, CatV$^2$TON \cite{chong2025catv2ton} and MagicTryOn \cite{li2025magictryon} adopt DiT-based backbones to unify spatial and temporal modeling for video virtual try-on.
Recent concurrent works further explore broader VVT settings, including dynamic camera trajectories \cite{sun2026tryoncrafter}, interactive human-garment manipulation \cite{zheng2026itryon}, unified fashion generation \cite{yang2026orthotryon}, and multi-object video try-on \cite{xia2026omnitryon}.

Although existing video virtual try-on methods have achieved promising results, most are developed on relatively low-resolution datasets, such as ViViD \cite{fang2024vivid} and VVT \cite{dong2019fw}. As a result, they mainly focus on coarse garment transfer and often struggle to preserve fine-grained details in high-resolution scenarios. Beyond resolution limitations, existing methods also inject garment visual features statically throughout denoising, leaving stage-adaptive fine-grained conditioning underexplored even on benchmarks such as Eevee \cite{zeng2025eeveecloseuphighresolutionvideobased}.
In contrast, our method dynamically modulates individual garment visual tokens based on their representations and the current denoising timestep, emphasizing coarse structural cues at early stages and progressively focusing on fine-grained textures and patterns at later stages.

\section{Method}
Our method is built upon a pretrained Diffusion Transformer for high-fidelity video virtual try-on.
Given a person video, clothing-agnostic masks, pose representations, and garment reference images, TexTailor generates try-on videos with temporal consistency and fine-grained garment details.
To effectively utilize these heterogeneous conditions, we introduce Timestep-Adaptive Visual Modulation (TAVM) for adaptive garment feature modulation, Frame-Aligned 3D Garment Cross-RoPE (FAC-RoPE) for spatial garment-video alignment, and Multi-Source Cross-Attention Injection (MCAI) for disentangled condition integration.
The overall framework of TexTailor is shown in Fig.~\ref{fig:overview}.

\subsection{Preliminary}
\label{Preliminary}
Our framework is built upon a pretrained latent video Diffusion Transformer trained with a Rectified Flow objective \cite{lipman2022flow}. Given a video clip \(V \in \mathbb{R}^{T \times H \times W \times 3}\), we first encode it into a latent representation:
\begin{equation}
z_0= \mathcal{E}(V) \in \mathbb{R}^{f \times c \times h \times w},
\end{equation}
where \(\mathcal{E}\) denotes the encoder of a variational autoencoder (VAE), and the corresponding decoder is denoted by \(\mathcal{D}\). Following the flow matching formulation, a latent trajectory is constructed by linearly interpolating between the clean latent \(z_0\) and a Gaussian prior sample \(\epsilon \sim \mathcal{N}(0, I)\):
\begin{equation}
z_t= (1-t) z_0 + t \epsilon, \qquad t \in [0,1].
\end{equation}
This interpolation defines a constant target velocity field
$u_t = \epsilon - z_0$.
A video DiT backbone \(\epsilon_{\theta}\) is trained to predict this target flow conditioned on the current latent \(z_t\), the timestep \(t\), and the conditioning information \(c\). The training objective is formulated as:
\begin{equation}
\mathcal{L}_{\mathrm{FM}} =
\mathbb{E}_{t, z_0, \epsilon}
\left[
w(t)\,
\bigl\|
\epsilon_{\theta}(z_t, t, c) - u_t
\bigr\|_2^2
\right],
\label{eq:FM}
\end{equation}
where \(w(t)\) is a timestep-dependent weighting function.

In our setting, the conditioning information \(c\) consists of two parts: a \emph{structural condition} and a \emph{garment condition}.
The structural condition includes the person video, pose video, agnostic video, and clothing mask, which together provide motion and spatial guidance for try-on generation. The garment condition is derived from available garment reference images, which are encoded into visual representations containing garment structure and fine-grained details.

\subsection{Timestep-Adaptive Visual Modulation}
Given garment reference images, we first extract complementary visual token sequences using SigLIP 2 and DINOv3 \cite{tschannen2025siglip,simeoni2025dinov3}.
We introduce Timestep-Adaptive Visual Modulation (TAVM) to provide adaptive garment visual guidance throughout the denoising process.
For each visual stream, TAVM applies timestep-conditioned AdaLN \cite{peebles2023scalable} to adapt visual features, producing modulated tokens \(\hat{X}=\{\hat{x}_i\}_{i=1}^{N}\).
A timestep-aware token-wise gate is then applied to reweight the modulated visual tokens:
\begin{equation}
g_i=\mathcal{G}(\hat{x}_i,e_t),
\end{equation}
where \(e_t\) denotes the timestep embedding, \(\mathcal{G}\) represents the gating function, and \(g_i\) indicates the contribution weight of the \(i\)-th visual token.
The visual tokens are then reweighted by the predicted gate weights:
\begin{equation}
\tilde{x}_i=g_i\cdot \hat{x}_i .
\end{equation}
By jointly conditioning feature modulation and token-wise gating on the denoising timestep, TAVM adaptively allocates garment visual information throughout generation, facilitating coarse-to-fine garment reconstruction.
The resulting SigLIP and DINO tokens are concatenated to form the visual condition, which is injected into the DiT backbone through multi-source cross-attention.

\subsection{Frame-Aligned 3D Garment Cross-RoPE}
Garment and video latents both preserve regular spatial grids, making it possible to explicitly model their spatial positional relationship.
In the native DiT backbone, each video token is encoded by 3D RoPE using its spatiotemporal coordinate $(f,h,w)$.
Garment latents are derived from a static reference image, preserving spatial positions $(h_g,w_g)$.
A straightforward extension to 3D is to assign them a fixed temporal coordinate, such as $f=0$.
For a video query at frame $f$, this creates an artificial temporal offset $f-0$ from the garment key, although the garment condition itself is static and shared across all frames.
To address this issue, we propose Frame-Aligned 3D Garment Cross-RoPE (FAC-RoPE).
For a video query at frame $f$, FAC-RoPE assigns the corresponding garment keys the same temporal coordinate $f$, extending their coordinates from $(h_g,w_g)$ to $(f,h_g,w_g)$.
The 3D RoPE is then applied as
\begin{align}
    \widehat{Q}_v^f&=\operatorname{RoPE}_{3D}(Q_v^f;f,h,w), \\
    \widehat{K}_g^f&=\operatorname{RoPE}_{3D}(K_g;f,h_g,w_g).
\end{align}
By assigning the same frame coordinate to the video query and garment key, FAC-RoPE preserves the native 3D spatiotemporal encoding of the video query while establishing frame-aligned positional correspondence with the garment key.
This design enables garment conditioning to exploit spatial correspondence without introducing artificial temporal positional bias, thereby improving local alignment and cross-frame texture consistency.

\subsection{Multi-Source Cross-Attention Injection}
Given the modulated visual tokens, we propose Multi-Source Cross-Attention Injection (MCAI) to integrate heterogeneous conditioning signals into the DiT backbone.
MCAI decomposes the conditioning space into three complementary sources: text semantics, modulated visual features, and garment latent tokens, which are injected through parallel cross-attention branches during the denoising process.
Each condition branch is independently projected and attended, allowing different modalities to preserve their own semantic characteristics while contributing complementary information to the generation process.
Compared with directly concatenating heterogeneous conditions into a shared token sequence, the disentangled multi-branch design reduces cross-modal interference and enables more effective utilization of garment-related cues.

\subsection{Training Objective}
In addition to the standard flow-matching objective, we introduce a \emph{mask-aware loss} to emphasize garment regions during training.
Since garment regions are critical to visual quality in video virtual try-on, treating all spatial locations equally may weaken supervision on clothing-related areas.
We therefore reweight the flow-matching objective using the mask video $M$ to provide stronger optimization signals for garment regions while maintaining the original generation objective.
Following Eq.~\ref{eq:FM}, the mask-aware term is defined as:
\begin{equation}
\mathcal{L}_{\mathrm{mask}}
=
\mathbb{E}_{t,z_0,\epsilon}
\left[
w(t)\left\|
M \odot
\left(
\epsilon_{\theta}(z_t,t,c)-u_t
\right)
\right\|_2^2
\right].
\end{equation}
The final objective is
\begin{equation}
\mathcal{L}
=
\mathcal{L}_{\mathrm{FM}}+\lambda \mathcal{L}_{\mathrm{mask}},
\end{equation}
where $\lambda$ controls the strength of mask-aware supervision.

\begin{figure*}[t]
    \centering
    \includegraphics[width=\textwidth]{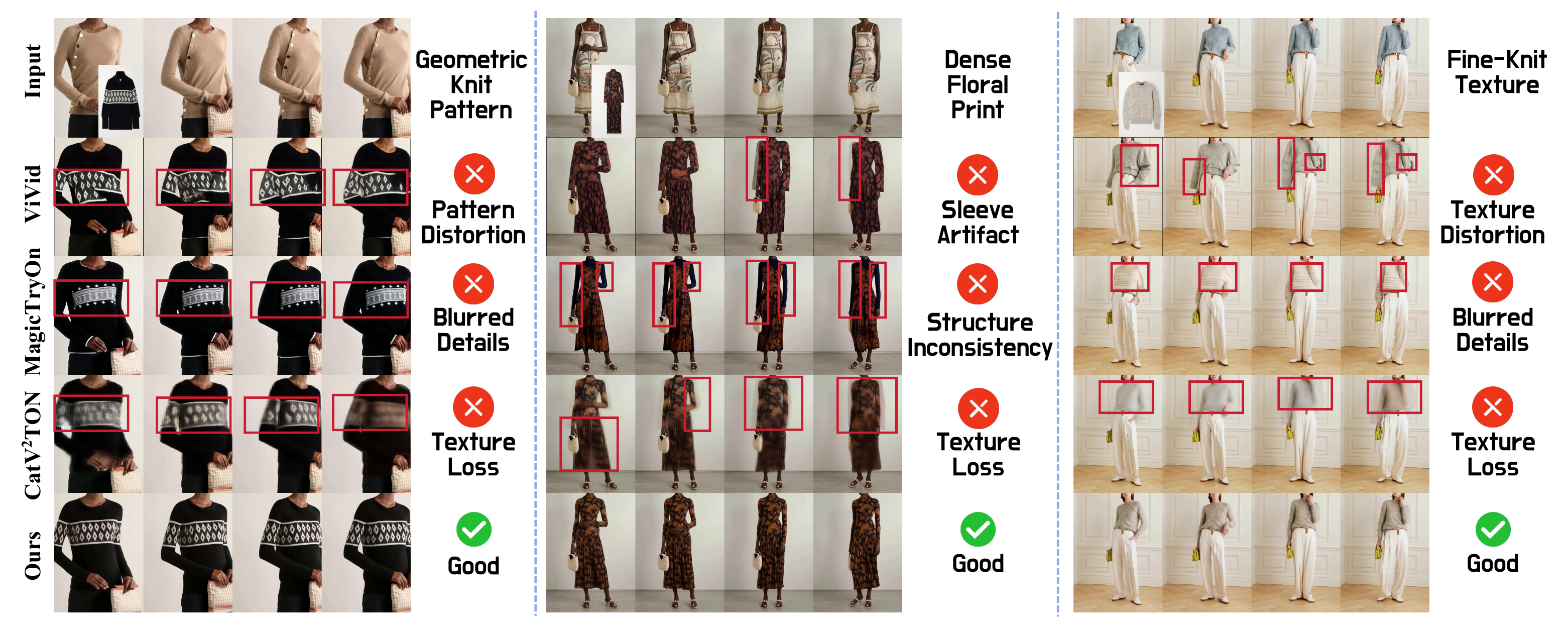}
    \caption{
    Qualitative comparison on the high-resolution Eevee benchmark. Compared with existing methods, TexTailor better preserves garment textures, patterns, and local structures while maintaining temporally coherent appearance across frames.
    }
    \label{fig:qualitative}
\end{figure*}

\begin{table*}[t]
\centering
{
\small
\setlength{\tabcolsep}{4pt}
\begin{tabular}{lcccccccc}
\toprule
\multirow{2}{*}{Method}
& \multicolumn{3}{c}{Full-shot}
& \multicolumn{3}{c}{Close-up}
& \multirow{2}{*}{GPU Mem.}
& \multirow{2}{*}{Time} \\
\cmidrule(lr){2-4}
\cmidrule(lr){5-7}
& VFID$\_R$ $\downarrow$
& VFID$\_I$ $\downarrow$
& VGID $\uparrow$
& VFID$\_R$ $\downarrow$
& VFID$\_I$ $\downarrow$
& VGID $\uparrow$
& & \\
\midrule

ViViD~\cite{fang2024vivid}
& 0.565
& 12.859
& 0.514
& 1.253
& 12.665
& 0.533
& 72.21G
& \underline{348.96s} \\

MagicTryOn~\cite{li2025magictryon}
& \underline{0.187}
& 9.783
& 0.512
& 0.752
& \underline{11.285}
& \underline{0.538}
& 69.54G
& 589.73s \\

CatV$^2$TON~\cite{chong2025catv2ton}
& 0.751
& \underline{9.086}
& \underline{0.520}
& \underline{0.673}
& 11.935
& 0.522
& \underline{40.18G}
& 357.21s \\

\textbf{TexTailor}
& \textbf{0.149}
& \textbf{8.872}
& \textbf{0.531}
& \textbf{0.489}
& \textbf{10.917}
& \textbf{0.543}
& \textbf{30.32G}
& \textbf{230.72s} \\

\bottomrule
\end{tabular}
}
\caption{
Quantitative comparison on the Eevee benchmark at ($1088 \times 816$).
We report results under both full-shot and close-up settings, together
with GPU memory usage and inference time. The best and second-best results are highlighted in \textbf{bold} and \underline{underline}, respectively.
}
\label{tab:eevee_highres}
\end{table*}

\begin{table*}[t]
\centering

{
\small
\setlength{\tabcolsep}{3.5pt}

\begin{tabular}{lcccccccc}
\toprule
Method
& VFID$_{I}^{p}$ $\downarrow$
& VFID$_{R}^{p}$ $\downarrow$
& SSIM $\uparrow$
& LPIPS $\downarrow$
& VFID$_{I}^{u}$ $\downarrow$
& VFID$_{R}^{u}$ $\downarrow$
& GPU Mem.
& Time \\
\midrule

ViViD~\cite{fang2024vivid}
& 17.1847
& 0.6382
& 0.8041
& 0.1216
& 21.6925
& 0.8347
& 62.59G
& \underline{204.183s} \\

CatV$^2$TON~\cite{chong2025catv2ton}
& 13.4821
& 0.2876
& 0.8742
& 0.0647
& 19.3846
& 0.5179
& \underline{27.66G}
& 209.127s \\

MagicTryOn~\cite{li2025magictryon}
& \underline{8.3165}
& \textbf{0.2298}
& \textbf{0.9024}
& \textbf{0.0598}
& \underline{14.6032}
& \underline{0.3137}
& 51.51G
& 345.271s \\

\textbf{TexTailor}
& \textbf{7.9428}
& \underline{0.2645}
& \underline{0.9017}
& \underline{0.0619}
& \textbf{13.8754}
& \textbf{0.2861}
& \textbf{25.32G}
& \textbf{196.852s} \\

\bottomrule
\end{tabular}
}

\caption{
Quantitative comparison on the ViViD benchmark. We report paired and
unpaired evaluation metrics, together with GPU memory usage and inference time. The best and second-best results are highlighted in \textbf{bold} and \underline{underline}, respectively.
}
\label{tab:vivid_benchmark}

\end{table*}

\begin{table*}[t]
\centering

{
\small
\setlength{\tabcolsep}{2.8pt}

\begin{tabular}{llccccccccc}
\toprule
Setting & Metric
& \shortstack{w/o\\TAVM-S}
& \shortstack{w/o\\TAVM-D}
& \shortstack{w/o\\TAVM-G}
& \shortstack{w/o\\TAVM}
& \shortstack{w/o\\FAC}
& \shortstack{w/o\\MCAI-T}
& \shortstack{w/o\\MCAI-G}
& \shortstack{w/o\\Mask}
& \shortstack{Full\\Model} \\
\midrule

\multirow{3}{*}{Full-shot}
& VFID$_R$ $\downarrow$
& 0.184 & 0.181 & 0.176 & 0.224 & 0.207
& 0.179 & 0.201 & 0.198 & \textbf{0.149} \\

& VFID$_I$ $\downarrow$
& 9.512 & 9.463 & 9.382 & 10.024 & 9.781
& 9.427 & 9.694 & 9.821 & \textbf{8.872} \\

& VGID $\uparrow$
& 0.520 & 0.521 & 0.522 & 0.512 & 0.515
& 0.523 & 0.516 & 0.517 & \textbf{0.531} \\

\midrule

\multirow{3}{*}{Close-up}
& VFID$_R$ $\downarrow$
& 0.551 & 0.543 & 0.535 & 0.612 & 0.581
& 0.529 & 0.568 & 0.592 & \textbf{0.489} \\

& VFID$_I$ $\downarrow$
& 11.612 & 11.534 & 11.462 & 12.146 & 11.873
& 11.421 & 11.754 & 11.982 & \textbf{10.917} \\

& VGID $\uparrow$
& 0.542 & 0.535 & 0.526 & 0.514 & 0.539
& 0.527 & 0.540 & 0.538 & \textbf{0.543} \\

\bottomrule
\end{tabular}
}
\caption{
Ablation study on the Eevee benchmark.
We evaluate the contribution of each component in TexTailor.
}

\label{tab:ablation_study}

\end{table*}

\section{Experiments}
\subsection{Experimental Setup}

\noindent\textbf{Datasets.}
We use the Eevee dataset \cite{zeng2025eeveecloseuphighresolutionvideobased} as the primary dataset for training and evaluation, as it provides high-resolution videos and diverse garment references suitable for evaluating fine-grained garment preservation. Compared with conventional video virtual try-on datasets such as ViViD \cite{fang2024vivid} and VVT \cite{dong2019fw}, Eevee provides videos at \(1088 \times 816\) resolution and garment images at \(2400 \times 1800\), enabling more comprehensive evaluation of garment appearance, textures, and local details. It covers three garment categories, including upper-body garments, lower-body garments, and dresses, with 4492, 2308, and 1564 training samples, and 500, 250, and 250 test samples, respectively. We report quantitative comparisons on both the Eevee dataset and the ViViD dataset to evaluate performance under both high-resolution and conventional video virtual try-on settings. More results under other evaluation settings are provided in the supplementary material.

\noindent\textbf{Evaluation settings and metrics.}
We evaluate our method under both paired and unpaired settings.
In the paired setting, the input garment is the same as the garment worn in the reference video.
In the unpaired setting, the input garment differs from the one originally worn by the person, which better reflects practical virtual try-on applications.
We report SSIM \cite{wang2004image}, LPIPS \cite{zhang2018unreasonable}, VFID-I3D \((\mathrm{VFID}_{I})\), VFID-ResNeXt \((\mathrm{VFID}_{R})\) \cite{unterthiner2018towards}, and VGID \cite{zeng2025eeveecloseuphighresolutionvideobased}.
SSIM and LPIPS measure reconstruction quality in the paired setting.
The two VFID variants, computed with I3D \cite{carreira2017quo} and ResNeXt \cite{xie2017aggregated} backbones, evaluate video fidelity and temporal consistency.
VGID measures garment-region semantic consistency using DINOv2 \cite{oquab2023dinov2} features, which provides an additional evaluation of garment identity preservation and fine-grained appearance consistency.
All metrics are reported in the paired setting, while only \(\mathrm{VFID}_{I}\), \(\mathrm{VFID}_{R}\), and VGID are reported in the unpaired setting.

\noindent\textbf{Implementation details.}
TexTailor is built upon the pretrained Wan2.2-Fun-5B-InP \cite{wan2025wan} video diffusion transformer.
During training, we apply LoRA \cite{hu2021lora} to the query projections of both self-attention and cross-attention layers in the DiT backbone.
The newly introduced modules, including TAVM, FAC-RoPE related projections, and MCAI branches, are fully fine-tuned to adapt the pretrained model for video virtual try-on.
Training is performed in two stages.
The model is first trained at a lower resolution for stable adaptation and then fine-tuned at \(1088 \times 816\) resolution to improve high-resolution garment detail preservation. Each training sample contains \(49\) video frames.
All experiments were conducted on 4 NVIDIA A100 (80GB) GPUs. 
During inference, all experiments use 25 denoising steps.
The same inference configuration is adopted for all compared methods to ensure a fair evaluation.

\subsection{Quantitative Comparison}

We evaluate TexTailor on both the Eevee and ViViD datasets and compare with representative video virtual try-on methods, including ViViD \cite{fang2024vivid}, MagicTryOn \cite{li2025magictryon}, and CatV$^2$TON \cite{chong2025catv2ton}. 
Quantitative results on the high-resolution Eevee setting at \(1088 \times 816\) and the ViViD dataset are reported in Table~\ref{tab:eevee_highres} and Table~\ref{tab:vivid_benchmark}, respectively.
These two benchmarks evaluate TexTailor under both high-resolution garment preservation and conventional video virtual try-on scenarios.
On the Eevee dataset, TexTailor achieves competitive performance under both full-shot and close-up settings.
The improvements in \(\mathrm{VFID}_{I}\), \(\mathrm{VFID}_{R}\), and VGID demonstrate that TexTailor effectively preserves fine-grained garment details in challenging high-resolution close-up scenarios, while maintaining high video fidelity and temporal consistency.
Results on the ViViD dataset further verify its generalization ability under conventional video virtual try-on settings.
Moreover, TexTailor achieves lower GPU memory consumption and faster inference speed compared with existing methods.

\subsection{Qualitative Comparison}
\label{sec:qualitative}
Fig.~\ref{fig:qualitative} presents qualitative comparisons between TexTailor and existing video virtual try-on methods. 
The examples cover diverse garment categories and challenging cases with complex textures, patterns, and local structures under various poses and viewpoints.
Compared with competing methods, TexTailor produces more faithful garment details and more coherent temporal appearance across different motion sequences.
Existing methods often preserve the coarse garment silhouette but struggle to maintain local details, where texture distortion, pattern blurring, and local misalignment become more visible in high-resolution scenarios with complex garment structures.
In contrast, TexTailor better retains fine-grained garment attributes such as collars and textures, while maintaining more stable appearance across frames.
This demonstrates that the proposed conditioning strategy effectively improves both spatial garment fidelity and temporal consistency in challenging video scenarios.
More qualitative comparisons are presented in the supplementary material.

\begin{figure*}[t]
    \centering
    \includegraphics[width=\textwidth]{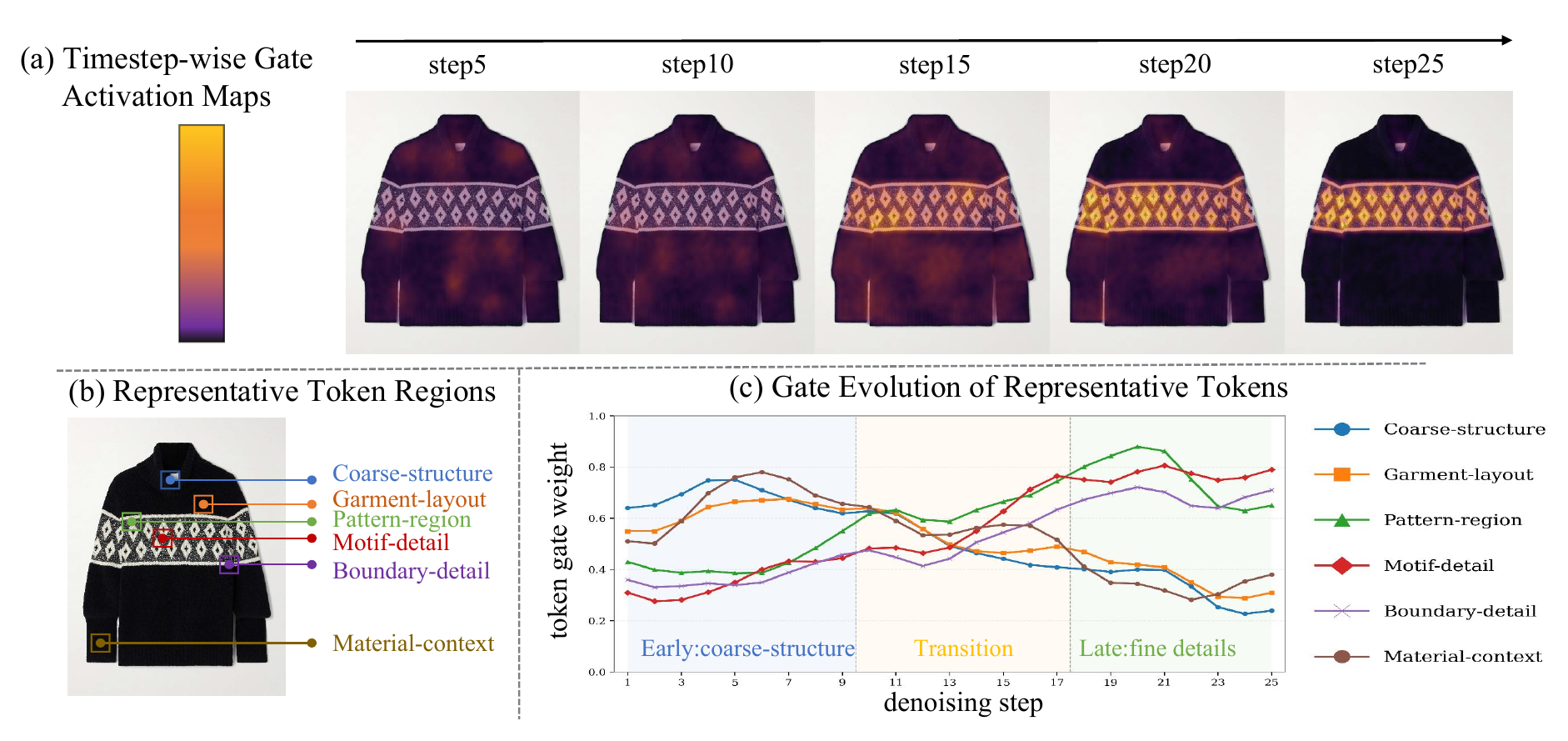}
    \caption{
    Visualization of timestep-wise token gating in TAVM.
    (a) Gate activation maps across different denoising steps.
    (b) Representative token regions selected from the garment reference.
    (c) Gate evolution curves of the selected token regions in (b).
    TAVM progressively shifts its emphasis from coarse garment structure to fine-grained local details during denoising.
    }
    \label{fig:tavm_visualization}
\end{figure*}
\subsection{Ablation Study}
\label{sec:ablation}
To validate the effectiveness of each proposed component, we conduct comprehensive ablation studies on the high-resolution Eevee dataset.
Quantitative results are reported in Table~\ref{tab:ablation_study}.
We investigate four key designs of TexTailor: TAVM, FAC-RoPE, MCAI, and the mask-aware loss.
Overall, removing any component leads to performance degradation, demonstrating that each design addresses a specific challenge in high-fidelity video virtual try-on.

\noindent\textbf{Effectiveness of TAVM.}
TAVM is designed to dynamically adjust garment visual guidance by leveraging complementary representations across different denoising stages.
To investigate the contribution of visual representations and timestep-aware modulation, we construct several variants.
Specifically, we remove individual visual streams to investigate the contribution of complementary garment representations, resulting in w/o TAVM-S and w/o TAVM-D for removing the SigLIP2 and DINOv3 branches, respectively. 
We further remove the timestep-aware token-wise gating mechanism (w/o TAVM-G) to evaluate the importance of adaptive visual selection during denoising. 
Finally, we remove the entire TAVM module to assess the overall contribution of timestep-adaptive visual modulation.
As shown in Table~\ref{tab:ablation_study}, removing either visual stream results in performance degradation, demonstrating that SigLIP2 and DINOv3 provide complementary visual representations.
Moreover, the degradation of w/o TAVM-G verifies that timestep-aware visual selection is crucial for progressively refining garment appearance during the denoising process.
To further understand how TAVM dynamically adjusts garment guidance during denoising, we visualize the timestep-wise token gating behavior in Fig.~\ref{fig:tavm_visualization}.
The gate activation maps reveal that TAVM gradually shifts its emphasis from coarse garment structures to fine-grained patterns and textures during denoising.
The token evolution curves further demonstrate that structure-related regions receive stronger activation in early stages, while detail-related regions gradually gain higher importance during later refinement stages.
These observations verify that TAVM performs stage-adaptive visual selection rather than uniformly injecting garment features throughout the denoising process.

\noindent\textbf{Effectiveness of FAC-RoPE.}
FAC-RoPE is introduced to establish explicit spatial correspondence between static garment representations and dynamic video latents.
To evaluate the importance of garment-video positional alignment, we remove the frame-aligned positional encoding while keeping the garment latent conditioning unchanged, denoted as w/o FAC.
As shown in Table~\ref{tab:ablation_study}, removing FAC-RoPE leads to noticeable performance degradation.
Without frame-aligned positional encoding, garment tokens lack consistent spatial anchors with video tokens, making it difficult to maintain accurate garment-region alignment during generation.
This demonstrates that explicit positional correspondence is essential for preserving local garment structures and improving temporal consistency across frames.

\noindent\textbf{Effectiveness of MCAI.}
MCAI is designed to reduce interference among heterogeneous conditioning signals by independently injecting different information sources through disentangled cross-attention branches.
To evaluate the contribution of different condition sources, we remove the text branch and garment latent branch, denoted as w/o MCAI-T and w/o MCAI-G, respectively.
As shown in Table~\ref{tab:ablation_study}, removing either branch leads to performance degradation, demonstrating that semantic guidance and spatially aligned garment features provide complementary information for high-fidelity try-on generation.

\noindent\textbf{Effectiveness of Mask-Aware Loss.}
Mask-aware loss is introduced to provide stronger supervision on garment regions, where visual fidelity is critical for video virtual try-on.
To evaluate its contribution, we remove the mask-aware loss during training, denoted as w/o mask.
As shown in Table~\ref{tab:ablation_study}, the performance degradation demonstrates that garment-region-focused supervision provides more effective optimization signals for reconstructing local textures, complex patterns, and subtle garment structures.

\section{Conclusion}
In this paper, we propose TexTailor, a high-fidelity video virtual try-on framework for fine-grained garment preservation in high-resolution video generation. 
TexTailor introduces timestep-adaptive visual modulation to dynamically utilize garment visual representations across denoising stages, frame-aligned 3D garment Cross-RoPE to establish stable garment-to-video positional correspondence, and multi-source cross-attention injection to reduce interference among heterogeneous conditions.
Extensive experiments on multiple video virtual try-on benchmarks, including the high-resolution Eevee benchmark, demonstrate that TexTailor better preserves local garment details.
These results highlight the importance of adaptive and spatially aligned garment conditioning for high-fidelity video virtual try-on. 
We hope TexTailor can provide a useful direction for future research on fine-grained garment modeling.

\bibliography{main}

@article{hu2026physeditworld,
  title={PhysEditWorld: A Large-Scale Dataset Toward Physics-Editable World Models},
  author={Hu, Bin and Ma, Yanwen and Huang, Jiehui and Zhang, Ziliang and Wu, Haoning and Zhang, Ruicheng and Li, Yaokun and Wang, Zijun and Zhang, Yuechen and Tseng, Chun-Mei and others},
  journal={arXiv preprint arXiv:2606.26694},
  year={2026}
}

@inproceedings{choi2024improving,
  title={Improving diffusion models for authentic virtual try-on in the wild},
  author={Choi, Yisol and Kwak, Sangkyung and Lee, Kyungmin and Choi, Hyungwon and Shin, Jinwoo},
  booktitle={European Conference on Computer Vision},
  pages={206--235},
  year={2024},
  organization={Springer}
}

@article{chong2024catvton,
  title={Catvton: Concatenation is all you need for virtual try-on with diffusion models},
  author={Chong, Zheng and Dong, Xiao and Li, Haoxiang and Zhang, Shiyue and Zhang, Wenqing and Zhang, Xujie and Zhao, Hanqing and Jiang, Dongmei and Liang, Xiaodan},
  journal={arXiv preprint arXiv:2407.15886},
  year={2024}
}

@article{wang2024stablegarment,
  title={Stablegarment: Garment-centric generation via stable diffusion},
  author={Wang, Rui and Guo, Hailong and Liu, Jiaming and Li, Huaxia and Zhao, Haibo and Tang, Xu and Hu, Yao and Tang, Hao and Li, Peipei},
  journal={arXiv preprint arXiv:2403.10783},
  year={2024}
}

@inproceedings{xu2025ootdiffusion,
  title={Ootdiffusion: Outfitting fusion based latent diffusion for controllable virtual try-on},
  author={Xu, Yuhao and Gu, Tao and Chen, Weifeng and Chen, Arlene},
  booktitle={Proceedings of the AAAI Conference on Artificial Intelligence},
  pages={8996--9004},
  year={2025}
}

@article{zhang2024mmtryon,
  title={Mmtryon: Multi-modal multi-reference control for high-quality fashion generation},
  author={Zhang, Xujie and Lin, Ente and Li, Xiu and Luo, Yuxuan and Kampffmeyer, Michael and Dong, Xin and Liang, Xiaodan},
  journal={arXiv preprint arXiv:2405.00448},
  year={2024}
}

@article{zheng2024viton,
  title={Viton-dit: Learning in-the-wild video try-on from human dance videos via diffusion transformers},
  author={Zheng, Jun and Zhao, Fuwei and Xu, Youjiang and Dong, Xin and Liang, Xiaodan},
  journal={arXiv preprint arXiv:2405.18326},
  year={2024}
}

@article{fang2024vivid,
  title={Vivid: Video virtual try-on using diffusion models},
  author={Fang, Zixun and Zhai, Wei and Su, Aimin and Song, Hongliang and Zhu, Kai and Wang, Mao and Chen, Yu and Liu, Zhiheng and Cao, Yang and Zha, Zheng-Jun},
  journal={arXiv preprint arXiv:2405.11794},
  year={2024}
}

@inproceedings{karras2024fashion,
  title={Fashion-vdm: Video diffusion model for virtual try-on},
  author={Karras, Johanna and Li, Yingwei and Liu, Nan and Zhu, Luyang and Yoo, Innfarn and Lugmayr, Andreas and Lee, Chris and Kemelmacher-Shlizerman, Ira},
  booktitle={SIGGRAPH Asia 2024 Conference Papers},
  pages={1--11},
  year={2024}
}

@article{chong2025catv2ton,
  title={Catv2ton: Taming diffusion transformers for vision-based virtual try-on with temporal concatenation},
  author={Chong, Zheng and Zhang, Wenqing and Zhang, Shiyue and Zheng, Jun and Dong, Xiao and Li, Haoxiang and Wu, Yiling and Jiang, Dongmei and Liang, Xiaodan},
  journal={arXiv preprint arXiv:2501.11325},
  year={2025}
}

@article{li2025magictryon,
  title={MagicTryOn: Harnessing Diffusion Transformer for Garment-Preserving Video Virtual Try-on},
  author={Li, Guangyuan and Zheng, Siming and Zhang, Hao and Chen, Jinwei and Luan, Junsheng and Ou, Binkai and Zhao, Lei and Li, Bo and Jiang, Peng-Tao},
  journal={arXiv preprint arXiv:2505.21325},
  year={2025}
}

@article{zuo2025dreamvvt,
  title={Dreamvvt: Mastering realistic video virtual try-on in the wild via a stage-wise diffusion transformer framework},
  author={Zuo, Tongchun and Huang, Zaiyu and Ning, Shuliang and Lin, Ente and Liang, Chao and Zheng, Zerong and Jiang, Jianwen and Zhang, Yuan and Gao, Mingyuan and Dong, Xin},
  journal={arXiv preprint arXiv:2508.02807},
  year={2025}
}

@misc{zeng2025eeveecloseuphighresolutionvideobased,
      title={Eevee: Towards Close-up High-resolution Video-based Virtual Try-on}, 
      author={Jianhao Zeng and Yancheng Bai and Ruidong Chen and Xuanpu Zhang and Lei Sun and Dongyang Jin and Ryan Xu and Nannan Zhang and Dan Song and Xiangxiang Chu},
      year={2025},
      eprint={2511.18957},
      archivePrefix={arXiv},
      primaryClass={cs.CV},
      url={https://arxiv.org/abs/2511.18957}, 
}

@inproceedings{dong2019fw,
  title={Fw-gan: Flow-navigated warping gan for video virtual try-on},
  author={Dong, Haoye and Liang, Xiaodan and Shen, Xiaohui and Wu, Bowen and Chen, Bing-Cheng and Yin, Jian},
  booktitle={Proceedings of the IEEE/CVF international conference on computer vision},
  pages={1161--1170},
  year={2019}
}

@article{chang2025pemf,
  title={PEMF-VTO: Point-Enhanced Video Virtual Try-on via Mask-free Paradigm},
  author={Chang, Tianyu and Chen, Xiaohao and Wei, Zhichao and Zhang, Xuanpu and Chen, Qingguo and Luo, Weihua and Song, Peipei and Yang, Xun},
  journal={IEEE Transactions on Consumer Electronics},
  year={2025},
  publisher={IEEE}
}

@article{li2025realvvt,
  title={Realvvt: Towards photorealistic video virtual try-on via spatio-temporal consistency},
  author={Li, Siqi and Jiang, Zhengkai and Zhou, Jiawei and Liu, Zhihong and Chi, Xiaowei and Wang, Haoqian},
  journal={arXiv preprint arXiv:2501.08682},
  year={2025}
}

@article{guo2023animatediff,
  title={Animatediff: Animate your personalized text-to-image diffusion models without specific tuning},
  author={Guo, Yuwei and Yang, Ceyuan and Rao, Anyi and Liang, Zhengyang and Wang, Yaohui and Qiao, Yu and Agrawala, Maneesh and Lin, Dahua and Dai, Bo},
  journal={arXiv preprint arXiv:2307.04725},
  year={2023}
}

@inproceedings{wu2023tune,
  title={Tune-a-video: One-shot tuning of image diffusion models for text-to-video generation},
  author={Wu, Jay Zhangjie and Ge, Yixiao and Wang, Xintao and Lei, Stan Weixian and Gu, Yuchao and Shi, Yufei and Hsu, Wynne and Shan, Ying and Qie, Xiaohu and Shou, Mike Zheng},
  booktitle={Proceedings of the IEEE/CVF international conference on computer vision},
  pages={7623--7633},
  year={2023}
}

@article{molad2023dreamix,
  title={Dreamix: Video diffusion models are general video editors},
  author={Molad, Eyal and Horwitz, Eliahu and Valevski, Dani and Acha, Alex Rav and Matias, Yossi and Pritch, Yael and Leviathan, Yaniv and Hoshen, Yedid},
  journal={arXiv preprint arXiv:2302.01329},
  year={2023}
}

@article{zhang2025mind,
  title={Mind-v: Hierarchical video generation for long-horizon robotic manipulation with rl-based physical alignment},
  author={Zhang, Ruicheng and Zhang, Mingyang and Zhou, Jun and Guo, Zhangrui and Liu, Xiaofan and Xu, Zunnan and Zhong, Zhizhou and Yan, Puxin and Luo, Haocheng and Li, Xiu},
  journal={arXiv e-prints},
  pages={arXiv--2512},
  year={2025}
}

@article{singer2022make,
  title={Make-a-video: Text-to-video generation without text-video data},
  author={Singer, Uriel and Polyak, Adam and Hayes, Thomas and Yin, Xi and An, Jie and Zhang, Songyang and Hu, Qiyuan and Yang, Harry and Ashual, Oron and Gafni, Oran and others},
  journal={arXiv preprint arXiv:2209.14792},
  year={2022}
}

@inproceedings{lai2025tracktention,
  title={Tracktention: Leveraging point tracking to attend videos faster and better},
  author={Lai, Zihang and Vedaldi, Andrea},
  booktitle={Proceedings of the Computer Vision and Pattern Recognition Conference},
  pages={22809--22819},
  year={2025}
}

@inproceedings{zhang2026zo3t,
  title={Zo3t: Zero-shot 3d-aware trajectory-guided image-to-video generation via test-time training},
  author={Zhang, Ruicheng and Zhou, Jun and Xu, Zunnan and Liu, Zihao and Huang, Jiehui and Zhang, Mingyang and Sun, Yu and Li, Xiu},
  booktitle={Proceedings of the AAAI Conference on Artificial Intelligence},
  volume={40},
  number={15},
  pages={12708--12716},
  year={2026}
}

@article{deng2025magref,
  title={MAGREF: Masked Guidance for Any-Reference Video Generation with Subject Disentanglement},
  author={Deng, Yufan and Yin, Yuanyang and Guo, Xun and Wang, Yizhi and Fang, Jacob Zhiyuan and Yuan, Shenghai and Yang, Yiding and Wang, Angtian and Liu, Bo and Huang, Haibin and others},
  journal={arXiv preprint arXiv:2505.23742},
  year={2025}
}

@article{chen2025dress,
  title={Dress\&Dance: Dress up and Dance as You Like It-Technical Preview},
  author={Chen, Jun-Kun and Bansal, Aayush and Vo, Minh Phuoc and Wang, Yu-Xiong},
  journal={arXiv preprint arXiv:2508.21070},
  year={2025}
}

@inproceedings{wei20253dv,
  title={3dv-ton: Textured 3d-guided consistent video try-on via diffusion models},
  author={Wei, Min and Yu, Chaohui and Zhou, Jingkai and Wang, Fan},
  booktitle={Proceedings of the 33rd ACM International Conference on Multimedia},
  pages={9345--9354},
  year={2025}
}

@article{kang2024mirror,
  title={Mirror: Towards generalizable on-device video virtual try-on for mobile shopping},
  author={Kang, Dong-Sig and Baek, Eunsu and Son, Sungwook and Lee, Youngki and Gong, Taesik and Kim, Hyung-Sin},
  journal={Proceedings of the ACM on Interactive, Mobile, Wearable and Ubiquitous Technologies},
  volume={7},
  number={4},
  pages={1--27},
  year={2024},
  publisher={ACM New York, NY, USA}
}

@inproceedings{dong2019towards,
  title={Towards multi-pose guided virtual try-on network},
  author={Dong, Haoye and Liang, Xiaodan and Shen, Xiaohui and Wang, Bochao and Lai, Hanjiang and Zhu, Jia and Hu, Zhiting and Yin, Jian},
  booktitle={Proceedings of the IEEE/CVF international conference on computer vision},
  pages={9026--9035},
  year={2019}
}

@inproceedings{he2024wildvidfit,
  title={Wildvidfit: Video virtual try-on in the wild via image-based controlled diffusion models},
  author={He, Zijian and Chen, Peixin and Wang, Guangrun and Li, Guanbin and Torr, Philip HS and Lin, Liang},
  booktitle={European Conference on Computer Vision},
  pages={123--139},
  year={2024},
  organization={Springer}
}

@inproceedings{peebles2023scalable,
  title={Scalable diffusion models with transformers},
  author={Peebles, William and Xie, Saining},
  booktitle={Proceedings of the IEEE/CVF international conference on computer vision},
  pages={4195--4205},
  year={2023}
}

@article{tschannen2025siglip,
  title={Siglip 2: Multilingual vision-language encoders with improved semantic understanding, localization, and dense features},
  author={Tschannen, Michael and Gritsenko, Alexey and Wang, Xiao and Naeem, Muhammad Ferjad and Alabdulmohsin, Ibrahim and Parthasarathy, Nikhil and Evans, Talfan and Beyer, Lucas and Xia, Ye and Mustafa, Basil and others},
  journal={arXiv preprint arXiv:2502.14786},
  year={2025}
}

@article{simeoni2025dinov3,
  title={Dinov3},
  author={Sim{\'e}oni, Oriane and Vo, Huy V and Seitzer, Maximilian and Baldassarre, Federico and Oquab, Maxime and Jose, Cijo and Khalidov, Vasil and Szafraniec, Marc and Yi, Seungeun and Ramamonjisoa, Micha{\"e}l and others},
  journal={arXiv preprint arXiv:2508.10104},
  year={2025}
}

@article{wang2004image,
  title={Image quality assessment: from error visibility to structural similarity},
  author={Wang, Zhou and Bovik, Alan C and Sheikh, Hamid R and Simoncelli, Eero P},
  journal={IEEE transactions on image processing},
  volume={13},
  number={4},
  pages={600--612},
  year={2004},
  publisher={IEEE}
}

@inproceedings{zhang2018unreasonable,
  title={The unreasonable effectiveness of deep features as a perceptual metric},
  author={Zhang, Richard and Isola, Phillip and Efros, Alexei A and Shechtman, Eli and Wang, Oliver},
  booktitle={Proceedings of the IEEE conference on computer vision and pattern recognition},
  pages={586--595},
  year={2018}
}

@article{unterthiner2018towards,
  title={Towards accurate generative models of video: A new metric \& challenges},
  author={Unterthiner, Thomas and Van Steenkiste, Sjoerd and Kurach, Karol and Marinier, Raphael and Michalski, Marcin and Gelly, Sylvain},
  journal={arXiv preprint arXiv:1812.01717},
  year={2018}
}

@article{oquab2023dinov2,
  title={Dinov2: Learning robust visual features without supervision},
  author={Oquab, Maxime and Darcet, Timoth{\'e}e and Moutakanni, Th{\'e}o and Vo, Huy and Szafraniec, Marc and Khalidov, Vasil and Fernandez, Pierre and Haziza, Daniel and Massa, Francisco and El-Nouby, Alaaeldin and others},
  journal={arXiv preprint arXiv:2304.07193},
  year={2023}
}

@inproceedings{carreira2017quo,
  title={Quo vadis, action recognition? a new model and the kinetics dataset},
  author={Carreira, Joao and Zisserman, Andrew},
  booktitle={proceedings of the IEEE Conference on Computer Vision and Pattern Recognition},
  pages={6299--6308},
  year={2017}
}

@inproceedings{xie2017aggregated,
  title={Aggregated residual transformations for deep neural networks},
  author={Xie, Saining and Girshick, Ross and Doll{\'a}r, Piotr and Tu, Zhuowen and He, Kaiming},
  booktitle={Proceedings of the IEEE conference on computer vision and pattern recognition},
  pages={1492--1500},
  year={2017}
}

@article{wan2025wan,
  title={Wan: Open and advanced large-scale video generative models},
  author={Wan, Team and Wang, Ang and Ai, Baole and Wen, Bin and Mao, Chaojie and Xie, Chen-Wei and Chen, Di and Yu, Feiwu and Zhao, Haiming and Yang, Jianxiao and others},
  journal={arXiv preprint arXiv:2503.20314},
  year={2025}
}

@article{hu2021lora,
  title={Lora: Low-rank adaptation of large language models},
  author={Hu, Edward J and Shen, Yelong and Wallis, Phillip and Allen-Zhu, Zeyuan and Li, Yuanzhi and Wang, Shean and Wang, Lu and Chen, Weizhu},
  journal={arXiv preprint arXiv:2106.09685},
  year={2021}
}

@article{pan2025once,
  title={Once Is Enough: Lightweight DiT-Based Video Virtual Try-On via One-Time Garment Appearance Injection},
  author={Pan, Yanjie and He, Qingdong and Wang, Lidong and Peng, Bo and Chi, Mingmin},
  journal={arXiv preprint arXiv:2510.07654},
  year={2025}
}

@inproceedings{li2025pursuing,
  title={Pursuing temporal-consistent video virtual try-on via dynamic pose interaction},
  author={Li, Dong and Zhong, Wenqi and Yu, Wei and Pan, Yingwei and Zhang, Dingwen and Yao, Ting and Han, Junwei and Mei, Tao},
  booktitle={Proceedings of the Computer Vision and Pattern Recognition Conference},
  pages={22648--22657},
  year={2025}
}

@inproceedings{zhou2025fireedit,
  title={Fireedit: Fine-grained instruction-based image editing via region-aware vision language model},
  author={Zhou, Jun and Li, Jiahao and Xu, Zunnan and Li, Hanhui and Cheng, Yiji and Hong, Fa-Ting and Lin, Qin and Lu, Qinglin and Liang, Xiaodan},
  booktitle={2025 IEEE/CVF Conference on Computer Vision and Pattern Recognition (CVPR)},
  pages={13093--13103},
  year={2025},
  organization={IEEE}
}

@article{zheng2024dynamic,
  title={Dynamic try-on: Taming video virtual try-on with dynamic attention mechanism},
  author={Zheng, Jun and Wang, Jing and Zhao, Fuwei and Zhang, Xujie and Liang, Xiaodan},
  journal={arXiv preprint arXiv:2412.09822},
  year={2024}
}

@inproceedings{zhu2023tryondiffusion,
  title={Tryondiffusion: A tale of two unets},
  author={Zhu, Luyang and Yang, Dawei and Zhu, Tyler and Reda, Fitsum and Chan, William and Saharia, Chitwan and Norouzi, Mohammad and Kemelmacher-Shlizerman, Ira},
  booktitle={Proceedings of the IEEE/CVF conference on computer vision and pattern recognition},
  pages={4606--4615},
  year={2023}
}

@inproceedings{gou2023taming,
  title={Taming the power of diffusion models for high-quality virtual try-on with appearance flow},
  author={Gou, Junhong and Sun, Siyu and Zhang, Jianfu and Si, Jianlou and Qian, Chen and Zhang, Liqing},
  booktitle={Proceedings of the 31st ACM international conference on multimedia},
  pages={7599--7607},
  year={2023}
}

@inproceedings{cui2025street,
  title={Street tryon: Learning in-the-wild virtual try-on from unpaired person images},
  author={Cui, Aiyu and Mahajan, Jay and Shah, Viraj and Gomathinayagam, Preeti and Liu, Chang and Lazebnik, Svetlana},
  booktitle={Proceedings of the Winter Conference on Applications of Computer Vision},
  pages={1414--1423},
  year={2025}
}

@inproceedings{kim2024stableviton,
  title={Stableviton: Learning semantic correspondence with latent diffusion model for virtual try-on},
  author={Kim, Jeongho and Gu, Guojung and Park, Minho and Park, Sunghyun and Choo, Jaegul},
  booktitle={Proceedings of the IEEE/CVF conference on computer vision and pattern recognition},
  pages={8176--8185},
  year={2024}
}

@inproceedings{shim2024towards,
  title={Towards squeezing-averse virtual try-on via sequential deformation},
  author={Shim, Sang-Heon and Chung, Jiwoo and Heo, Jae-Pil},
  booktitle={Proceedings of the AAAI Conference on Artificial Intelligence},
  pages={4856--4863},
  year={2024}
}

@article{zhang2026kvpo,
  title={Kvpo: Ode-native grpo for autoregressive video alignment via kv semantic exploration},
  author={Zhang, Ruicheng and Cong, Kaixi and Zhou, Jun and Zhong, Zhizhou and Xu, Zunnan and Mao, Shuiyang and Liu, Wei and Li, Xiu},
  journal={arXiv preprint arXiv:2605.14278},
  year={2026}
}

@inproceedings{hong2025audio,
  title={Audio-visual controlled video diffusion with masked selective state spaces modeling for natural talking head generation},
  author={Hong, Fa-Ting and Xu, Zunnan and Zhou, Zixiang and Zhou, Jun and Li, Xiu and Lin, Qin and Lu, Qinglin and Xu, Dan},
  booktitle={2025 IEEE/CVF International Conference on Computer Vision (ICCV)},
  pages={12549--12558},
  year={2025},
  organization={IEEE}
}

@inproceedings{yang2025omnivton,
  title={Omnivton: Training-free universal virtual try-on},
  author={Yang, Zhaotong and Li, Yuhui and He, Shengfeng and Li, Xinzhe and Xu, Yangyang and Dong, Junyu and Du, Yong},
  booktitle={Proceedings of the IEEE/CVF International Conference on Computer Vision},
  pages={16702--16711},
  year={2025}
}

@inproceedings{guo2025any2anytryon,
  title={Any2anytryon: Leveraging adaptive position embeddings for versatile virtual clothing tasks},
  author={Guo, Hailong and Zeng, Bohan and Song, Yiren and Zhang, Wentao and Liu, Jiaming and Zhang, Chuang},
  booktitle={Proceedings of the IEEE/CVF International Conference on Computer Vision},
  pages={19085--19096},
  year={2025}
}

@inproceedings{zhou2025learning,
  title={Learning flow fields in attention for controllable person image generation},
  author={Zhou, Zijian and Liu, Shikun and Han, Xiao and Liu, Haozhe and Ng, Kam Woh and Xie, Tian and Cong, Yuren and Li, Hang and Xu, Mengmeng and P{\'e}rez-R{\'u}a, Juan-Manuel and others},
  booktitle={Proceedings of the Computer Vision and Pattern Recognition Conference},
  pages={2491--2501},
  year={2025}
}

@article{sun2026tryoncrafter,
  title={TryOnCrafter: Unleashing Camera Trajectories for Realistic Video Virtual Try-on via a Renderable 4D Try-on Proxy},
  author={Sun, Hao and Yan, Hao and Chen, Mengting and Song, Quanjian and Li, Yu and Cao, Juan and Lan, Jinsong and Zhu, Xiaoyong and Zheng, Bo and Tang, Sheng},
  journal={arXiv preprint arXiv:2606.26092},
  year={2026}
}

@article{yang2026orthotryon,
  title={OrthoTryOn: Geometric Orthogonalization for Conflict-Free Unified Fashion Generation},
  author={Yang, Zhaotong and Tai, Ying and Zhan, Jiahui and Zheng, Yu and Qian, Jianjun and Yang, Jian},
  journal={arXiv preprint arXiv:2606.27880},
  year={2026}
}

@article{zheng2026itryon,
  title={iTryOn: Mastering Interactive Video Virtual Try-On with Spatial-Semantic Guidance},
  author={Zheng, Jun and Xu, Zhengze and Chen, Mengting and Wang, Jing and Lan, Jinsong and Zhu, Xiaoyong and Zhang, Kaifu and Zheng, Bo and Liang, Xiaodan},
  journal={arXiv preprint arXiv:2605.21431},
  year={2026}
}

@article{xia2026omnitryon,
  title={OmniTryOn: Video Try-On Anything at Once!},
  author={Xia, Changliang and Jia, Chengyou and Luo, Minnan and Dang, Zhuohang and Shen, Xin and Ping, Bowen},
  journal={arXiv preprint arXiv:2606.08514},
  year={2026}
}

@article{sai2026modaflow,
  title={ModaFlow: Modality-Aware Flow Matching for High-Fidelity Virtual Try-On},
  author={Sai, Xiangyu and Madadi, Meysam and Escalera, Sergio and Xu, Yong},
  journal={arXiv preprint arXiv:2606.27773},
  year={2026}
}

@article{lipman2022flow,
  title={Flow matching for generative modeling},
  author={Lipman, Yaron and Chen, Ricky TQ and Ben-Hamu, Heli and Nickel, Maximilian and Le, Matt},
  journal={arXiv preprint arXiv:2210.02747},
  year={2022}
}

@article{huang2026consistentid,
  title={Consistentid: Portrait generation with multimodal fine-grained identity preserving},
  author={Huang, Jiehui and Dong, Xiao and Song, Wenhui and Chong, Zheng and Tang, Zhenchao and Zhou, Jun and Cheng, Yuhao and Chen, Long and Li, Hanhui and Yan, Yiqiang and others},
  journal={IEEE Transactions on Pattern Analysis and Machine Intelligence},
  year={2026},
  publisher={IEEE}
}

@article{hu2026embedding,
  title={Embedding-perturbed exploration preference optimization for flow models},
  author={Hu, Sujie and Chen, Chubin and Zhu, Jiashu and Wu, Jiahong and Chu, Xiangxiang and Li, Xiu},
  journal={arXiv preprint arXiv:2605.15803},
  year={2026}
}

@article{liu2025controllable,
  title={Controllable layer decomposition for reversible multi-layer image generation},
  author={Liu, Zihao and Xu, Zunnan and Shu, Shi and Zhou, Jun and Zhang, Ruicheng and Tang, Zhenchao and Li, Xiu},
  journal={arXiv preprint arXiv:2511.16249},
  year={2025}
}

@article{kong2024hunyuanvideo,
  title={Hunyuanvideo: A systematic framework for large video generative models},
  author={Kong, Weijie and Tian, Qi and Zhang, Zijian and Min, Rox and Dai, Zuozhuo and Zhou, Jin and Xiong, Jiangfeng and Li, Xin and Wu, Bo and Zhang, Jianwei and others},
  journal={arXiv preprint arXiv:2412.03603},
  year={2024}
}

@inproceedings{xu2025hunyuanportrait,
  title={Hunyuanportrait: Implicit condition control for enhanced portrait animation},
  author={Xu, Zunnan and Yu, Zhentao and Zhou, Zixiang and Zhou, Jun and Jin, Xiaoyu and Hong, Fa-Ting and Ji, Xiaozhong and Zhu, Junwei and Cai, Chengfei and Tang, Shiyu and others},
  booktitle={2025 IEEE/CVF Conference on Computer Vision and Pattern Recognition (CVPR)},
  pages={15909--15919},
  year={2025},
  organization={IEEE}
}

@article{hu2026identity,
  title={Identity-Consistent Video Generation under Large Facial-Angle Variations},
  author={Hu, Bin and Qi, Zipeng and Huang, Guoxi and Xu, Zunnan and Zhang, Ruicheng and Ye, Chongjie and Zhou, Jun and Li, Xiu and Wang, Jingdong},
  journal={arXiv preprint arXiv:2603.21299},
  year={2026}
}

\clearpage
\appendix
\section{Supplementary Materials}

This supplementary material provides additional experimental results and analysis to complement the main paper.
We first present more quantitative results under different resolutions, followed by additional qualitative comparisons and visualization results.
We then provide further ablation analysis and discuss the limitations of the current framework.

\subsection{More Quantitative Results}

\begin{table*}[t]
\centering

{
\small
\setlength{\tabcolsep}{4pt}

\begin{tabular}{lcccccccc}
\toprule
\multirow{2}{*}{Method}
& \multicolumn{3}{c}{Full-shot}
& \multicolumn{3}{c}{Close-up}
& \multirow{2}{*}{GPU Mem.}
& \multirow{2}{*}{Time} \\
\cmidrule(lr){2-4}
\cmidrule(lr){5-7}

& VFID$_R$ $\downarrow$
& VFID$_I$ $\downarrow$
& VGID $\uparrow$
& VFID$_R$ $\downarrow$
& VFID$_I$ $\downarrow$
& VGID $\uparrow$
& & \\
\midrule

ViViD~\cite{fang2024vivid}
& 0.389 & 12.194 & 0.506
& 0.936 & 12.198 & 0.533
& 64.12G & 216.734s \\

MagicTryOn~\cite{li2025magictryon}
& \underline{0.161} & 9.865 & \underline{0.520}
& \underline{0.595} & \underline{11.262} & 0.534
& 49.83G & 333.906s \\

CatV$^2$TON~\cite{chong2025catv2ton}
& 0.746 & \underline{9.141} & 0.518
& 0.632 & 11.847 & \underline{0.538}
& \underline{29.14G} & \underline{197.683s} \\

\textbf{TexTailor}
& \textbf{0.153}
& \textbf{8.996}
& \textbf{0.529}
& \textbf{0.497}
& \textbf{11.071}
& \textbf{0.553}
& \textbf{24.87G}
& \textbf{188.436s} \\

\bottomrule
\end{tabular}
}

\caption{
Quantitative comparison on the Eevee benchmark at \(832 \times 624\).
We report results under both full-shot and close-up settings, together
with GPU memory usage and inference time. The best and second-best results
are highlighted in \textbf{bold} and \underline{underline}, respectively.
}

\label{tab:quantitative_lowres}

\end{table*}

We provide additional quantitative results on the Eevee dataset at  \(832 \times 624\) resolution in Table~\ref{tab:quantitative_lowres}. 
TexTailor maintains competitive performance under both full-shot and close-up settings. 
Moreover, our method achieves lower GPU memory consumption and faster inference speed than existing approaches.

\subsection{More Qualitative Results}
\begin{figure*}[t]
    \centering
    \includegraphics[width=\textwidth]{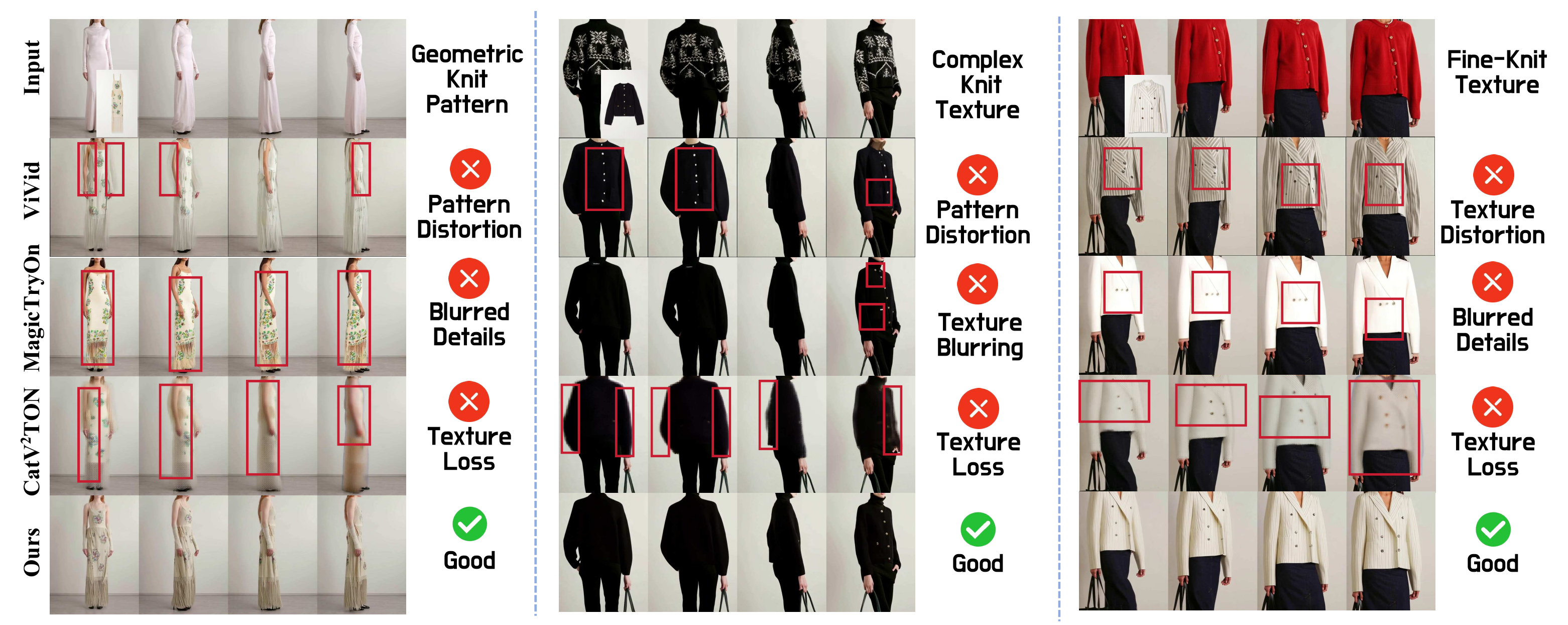}
    \caption{
  Qualitative comparison on challenging garment cases.
TexTailor better preserves complex patterns, fine-grained textures, and local garment structures compared with existing methods.
    }
    \label{fig:qualitative_append1}
\end{figure*}

\begin{figure*}[t]
    \centering
    \includegraphics[width=\textwidth]{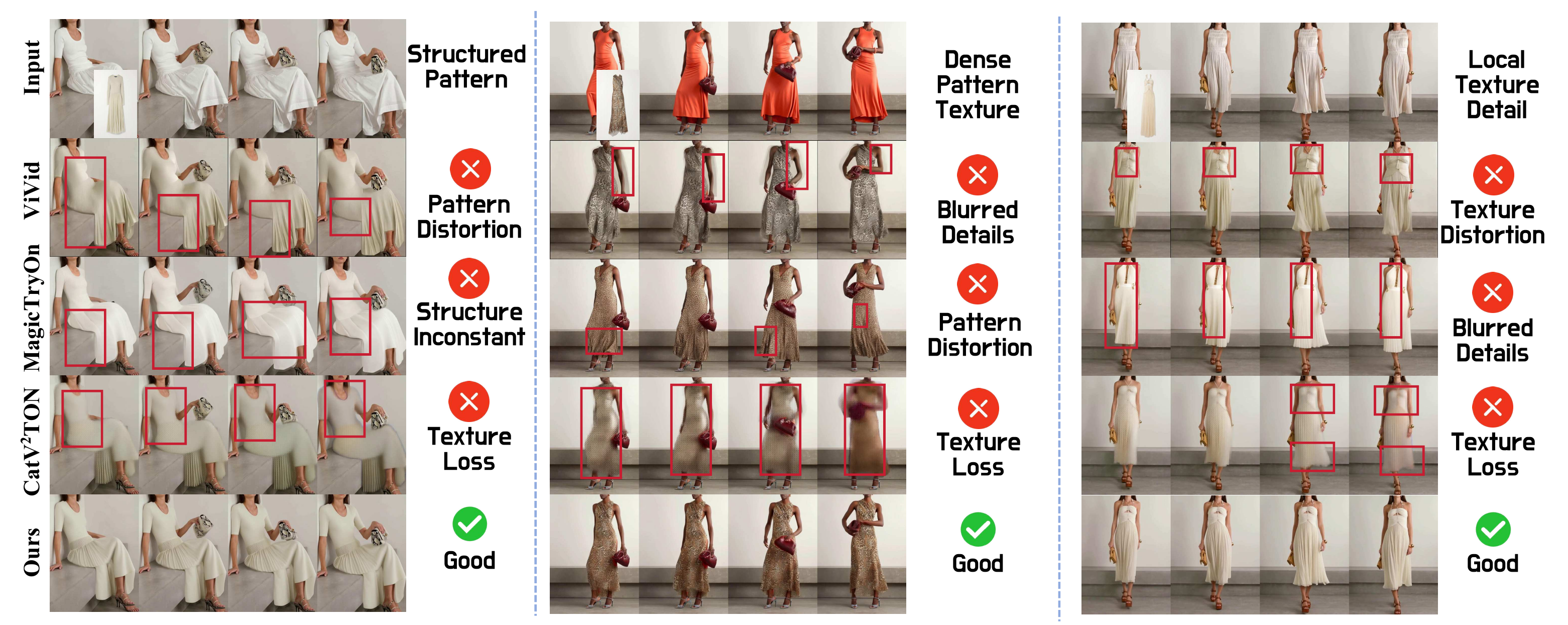}
\caption{
Qualitative comparison across diverse garment categories.
TexTailor achieves more faithful garment transfer with improved texture preservation and structural consistency.
}
    \label{fig:qualitative_append2}
\end{figure*}

\begin{figure*}[t]
    \centering
    \includegraphics[width=\textwidth]{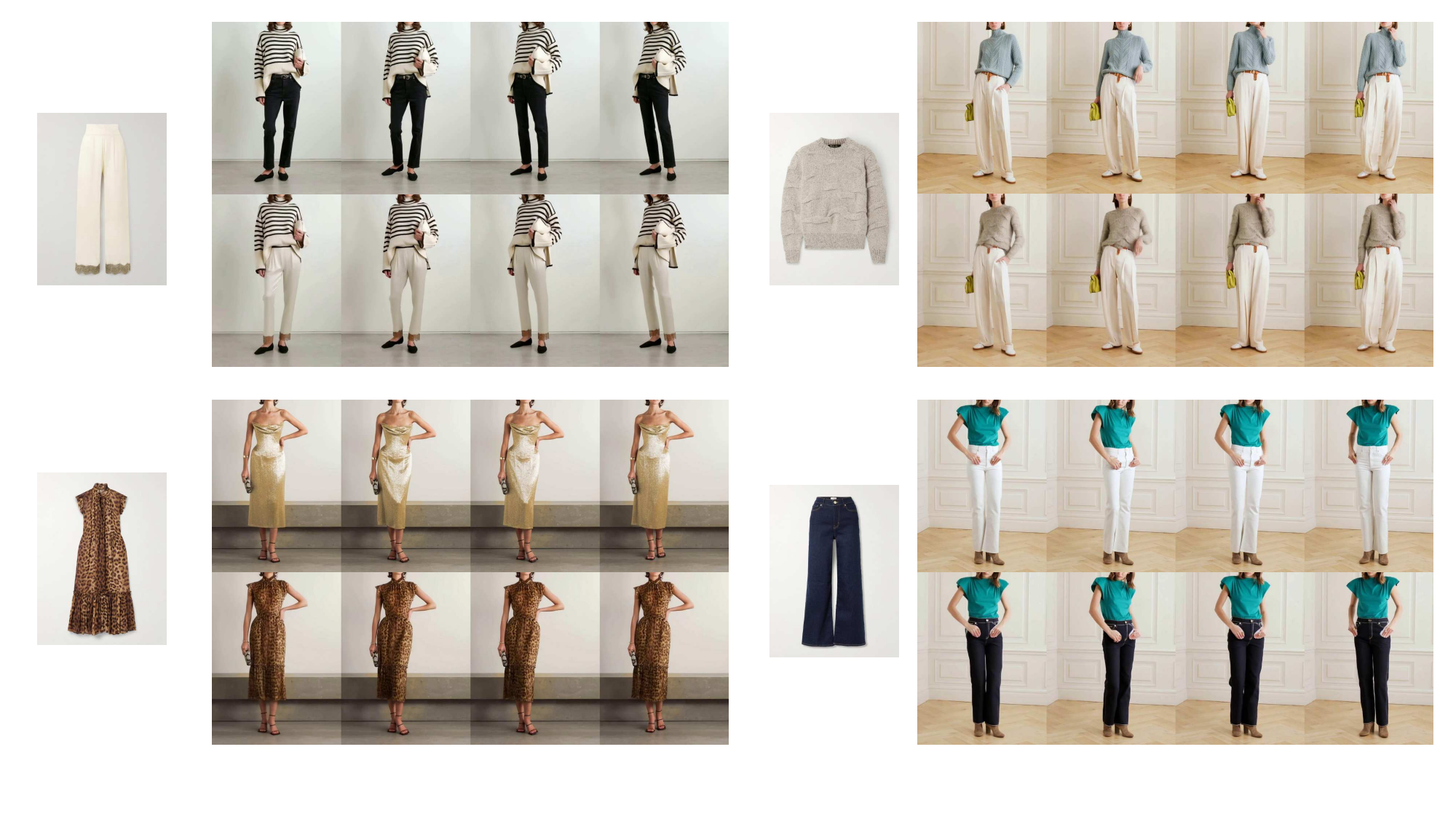}
\caption{
Additional results of TexTailor on diverse garment categories, including sweaters, dresses, and pants.
}
    \label{fig:qualitative_append3}
\end{figure*}

\begin{figure*}[t]
    \centering
    \includegraphics[width=\textwidth]{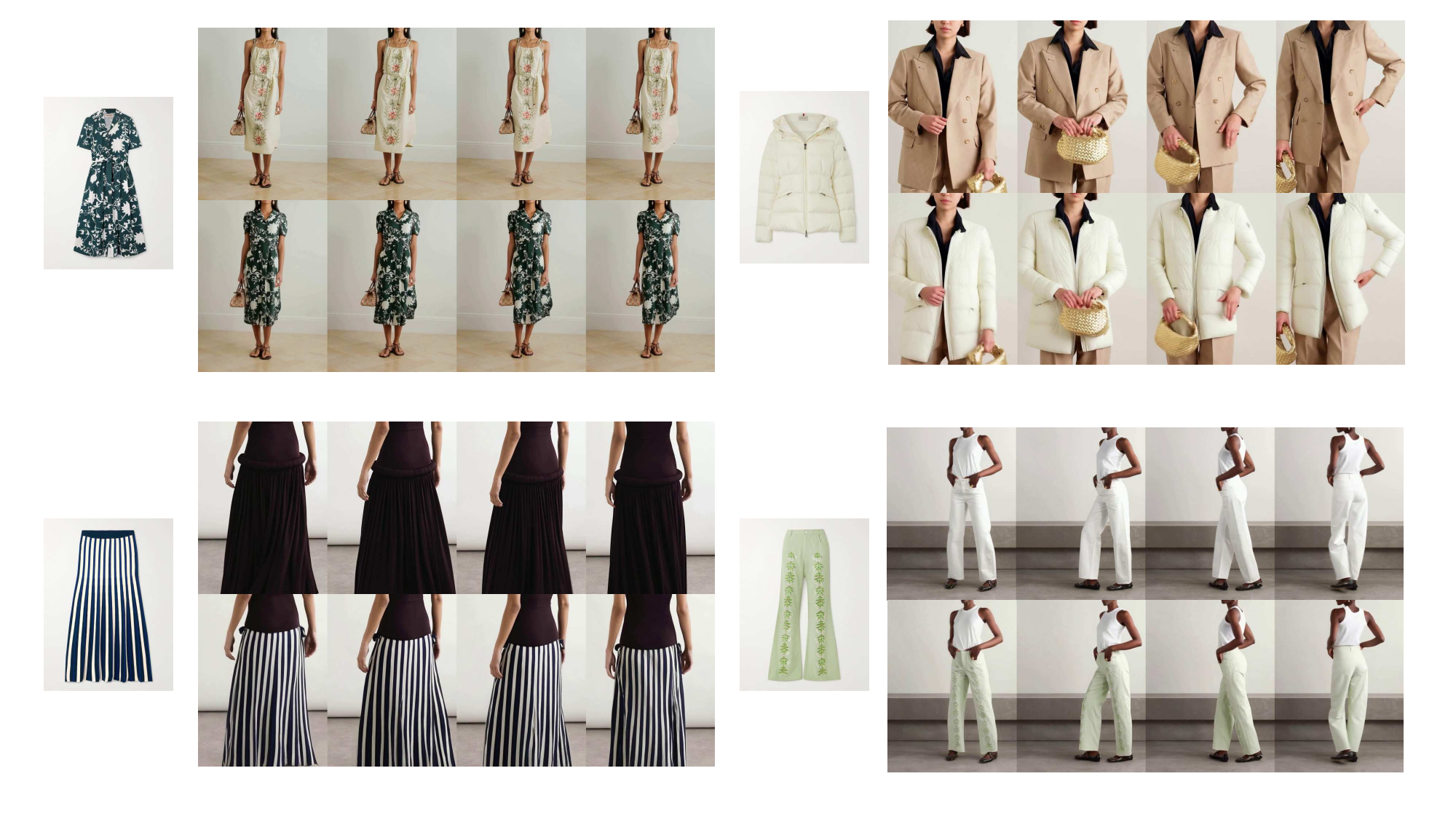}
\caption{
Additional results of TexTailor on garments with complex patterns and structures.
}
    \label{fig:qualitative_append4}
\end{figure*}

\begin{figure*}[t]
    \centering
    \includegraphics[width=\textwidth]{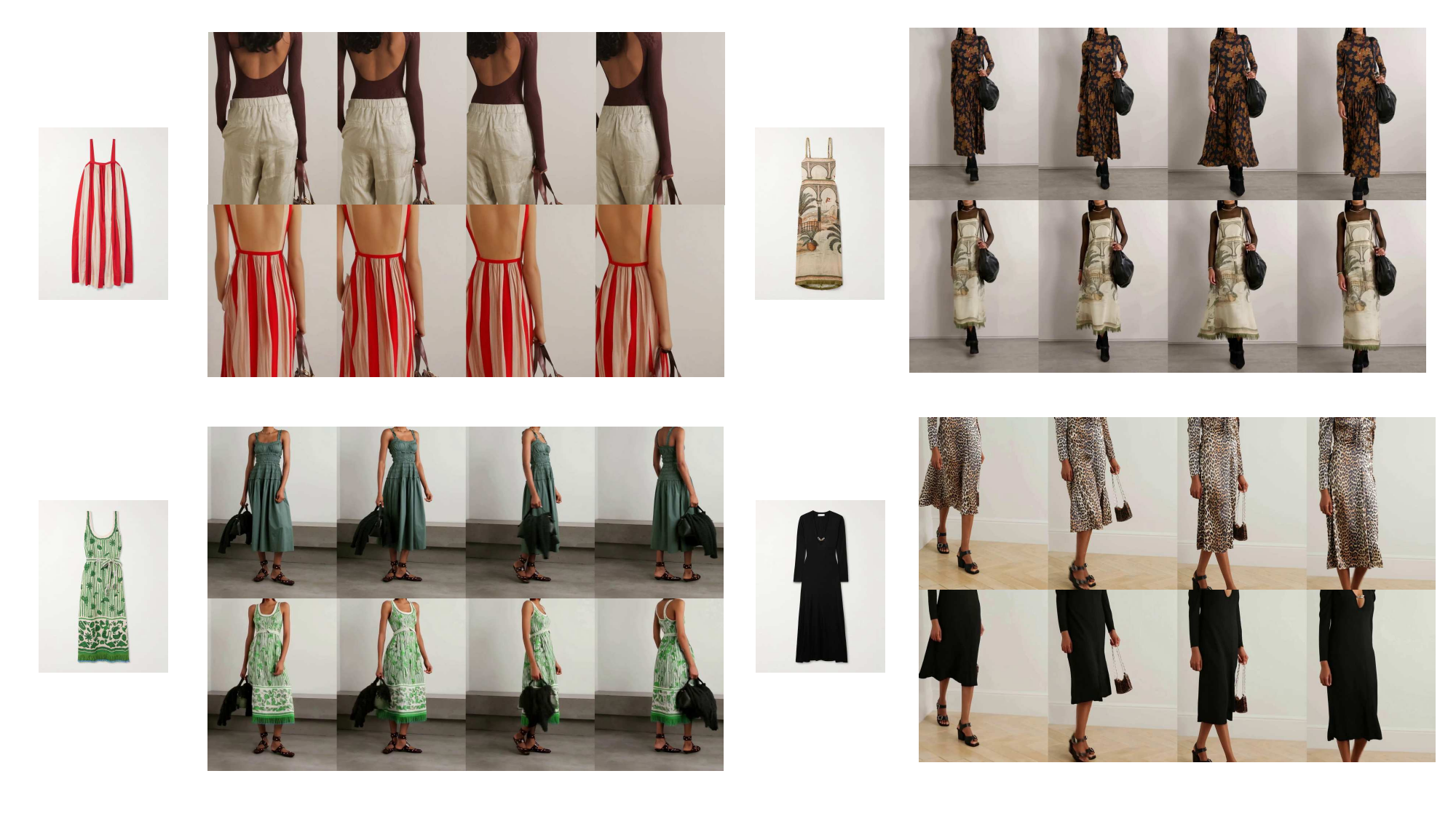}
\caption{
Additional results demonstrating the generalization ability of TexTailor across various garments and motion scenarios.
}
    \label{fig:qualitative_append5}
\end{figure*}

We provide additional qualitative results in Figs.~\ref{fig:qualitative_append1}-\ref{fig:qualitative_append5}.
These examples further demonstrate the effectiveness of TexTailor in preserving garment appearance, fine-grained details, and temporal consistency across diverse garment categories and video scenarios.

\subsection{More Ablation Results}

\begin{figure*}[t]
    \centering
    \includegraphics[width=\textwidth]{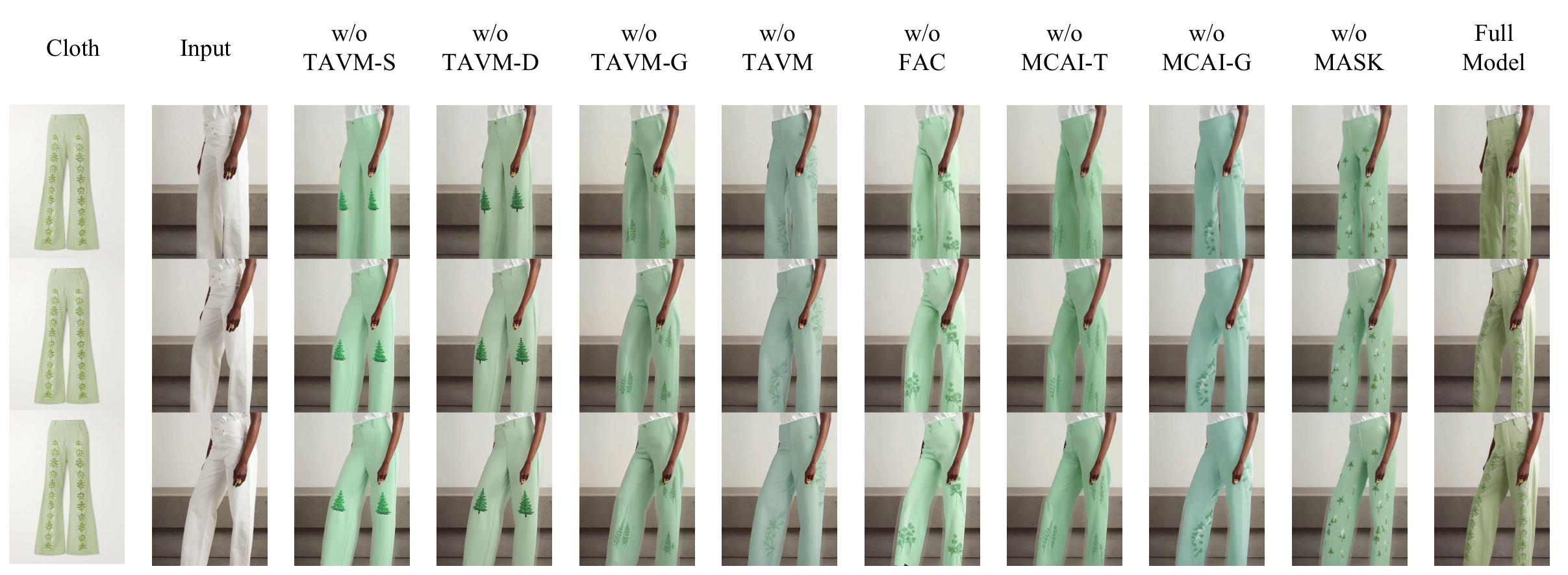}
    \caption{
    Qualitative ablation comparison of TexTailor.
    Removing individual components leads to degraded garment fidelity, including texture distortion, blurred details, and weaker pattern preservation.
    The full model achieves the most faithful garment appearance.
    }
    \label{fig:ablation}
\end{figure*}

We provide additional qualitative ablation results in Fig.~\ref{fig:ablation}. Removing individual components leads to visible degradation in garment fidelity, including texture distortion, blurred local details, and weaker pattern preservation.
The full model achieves more faithful garment reconstruction by effectively combining adaptive visual modulation, frame-aligned garment conditioning, multi-source condition injection, and mask-aware supervision. These qualitative results are consistent with the quantitative ablation results, further validating the effectiveness of each component in TexTailor.

\subsection{Discussion and Limitations}
\label{app:discuss}

Although TexTailor achieves strong performance in garment preservation and temporal consistency, several limitations remain. 
First, our framework still relies on auxiliary inputs, including pose information, agnostic representations, and video masks, resulting in a relatively complex preprocessing pipeline. Developing a fully mask-free video virtual try-on framework with minimal external guidance remains an interesting direction for future research.
Second, challenging scenarios involving large deformation, heavy occlusion, or significant viewpoint changes remain difficult due to the inherent complexity of video generation.
Future work will explore more efficient and simplified conditioning pipelines, as well as improve robustness under complex motion and viewpoint variations for high-resolution video virtual try-on.

\end{document}